\documentclass[10pt,twocolumn,letterpaper]{article}

\usepackage{cvpr}              % To produce the CAMERA-READY version
\usepackage{graphicx}
\usepackage{amsmath}
\usepackage{amssymb}
\usepackage{booktabs}

\usepackage{microtype}      % microtypography
\usepackage{xcolor}         % colors
\usepackage{enumitem}
\usepackage{graphicx}
\usepackage{caption}
\usepackage{subcaption}
\usepackage{xcolor}
\usepackage{multirow}
\usepackage{amsmath}
\usepackage{wrapfig}
\usepackage{colortbl}

\usepackage[normalem]{ulem}

\usepackage[pagebackref,breaklinks,colorlinks]{hyperref}

\usepackage[capitalize]{cleveref}
\crefname{section}{Sec.}{Secs.}
\Crefname{section}{Section}{Sections}
\Crefname{table}{Table}{Tables}
\crefname{table}{Tab.}{Tabs.}

\def\cvprPaperID{1447} % *** Enter the CVPR Paper ID here
\def\confName{CVPR}
\def\confYear{2023}

\begin{document}

\title{Open Vocabulary Object Detection with \\ Proposal Mining and Prediction Equalization}

% \and

\author{Peixian Chen$^\dag$\footnotemark[1], ~~
Kekai Sheng$^\dag$\footnotemark[1], ~~
Mengdan Zhang$^\dag$\footnotemark[4], ~~
Mingbao Lin$\dag$, ~~ \\
Yunhang Shen$^\dag$, ~~ 
Shaohui Lin$^\ddag$, ~~
Bo Ren$^\dag$, ~~
Ke Li$^\dag$, ~~ \\
[0.2cm]
$^\dag$ Tencent Youtu Lab  ~~~ 
$^\ddag $ East China Normal University
}

\maketitle

\begin{abstract}
Although learning from a pre-trained vision-language model is efficacious for open-vocabulary object detection~(OVD) that identifies objects beyond the training vocabulary, two issues remain open, including proposal-level vision-language alignment and base-novel category prediction balance.
% 
% In this paper, we introduce MEDet to alleviate these issues.
In this paper, we introduce a novel proposal \textbf{M}ining and prediction \textbf{E}qualization framework for open-vocabulary object \textbf{Det}ection~(MEDet) to alleviate these issues.
Specifically, we perform proposal mining by refining the inherited vision-semantic knowledge in a coarse-to-fine and online manner, allowing for detection-oriented proposal-level feature alignment.
Meanwhile, we equalize prediction confidence by reinforcing novel category predictions with an offline class-wise adjustment, permitting the overall OVD performance gains.
% %
% In this paper, we introduce MEDet with its two contributions in proposal mining and prediction equalization.
% %
% The former is achieved by online refining the inherited vision-semantic knowledge in a coarse-to-fine manner, allowing for proposal-level detection-oriented feature alignment.
% %
% The latter is realized by reinforcing novel category predictions in an offline class-wise adjustment, permitting the overall OVD performance gains.
%
Extensive experiments demonstrate the superiority of MEDet over the state-of-the-art methods.
In particular, we increase $m$AP of novel categories from $29.1\%$ to $32.6\%$ on MS COCO and obtain $22.4\%$ mask AP on LVIS with gains of $1.4\%$.
% ($1.4\%$ increase)
% 
% In particular, we obtain {\color{red}3.5\%} AP$50$ increase on MS COCO and {\color{red}1.4\%} mask AP increase on LVIS over the competing approaches in detecting novel categories.
%
%Code is anonymously available at \url{https://github.com/Anonymous-px/MEDet}.
For the sake of reproducibility, code is anonymously released~\footnote{\url{https://github.com/peixianchen/MEDet}}.

\end{abstract}

\renewcommand{\thefootnote}{\fnsymbol{footnote}}
\footnotetext[1]{These authors contributed equally to this work.}
\footnotetext[4]{Corresponding author.}
\section{Introduction}
\label{Sec:intro}
% 开集检测setting介绍：
%
Developing object detection is one of the most elementary missions in computer vision. Though efforts have been made~\cite{CascadeRCNN,centernet,fasterrcnn}, the remarkable success mostly relies on the permit of accessing fully annotated datasets. However, annotating datasets requires a time laborious and lavish process, which barricades their scalability in category size and further prevents their applicability in real-world applications.
Fortunately, the recent prosperity of pre-trained vision-language models~(VLMs)~\cite{clip} opens a new horizon for open-vocabulary object detection~(OVD) that expands the detection vocabulary beyond the training categories.

\begin{figure}[!t]
    \centering
    \begin{subfigure}[b]{0.43\textwidth}
         \centering
         \includegraphics[width=\linewidth]{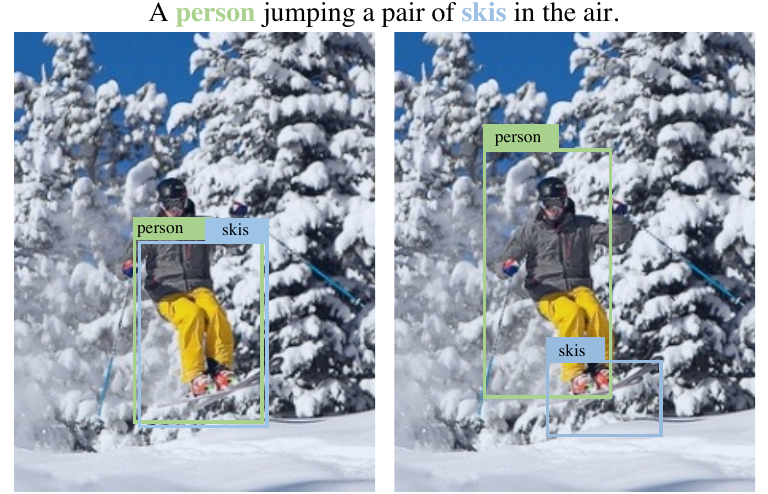}
         \caption{}
         \label{fig:proposal_problem}
    \end{subfigure}
    % \hfill
    \begin{subfigure}[b]{0.43\textwidth}
         \centering
         \includegraphics[width=\linewidth]{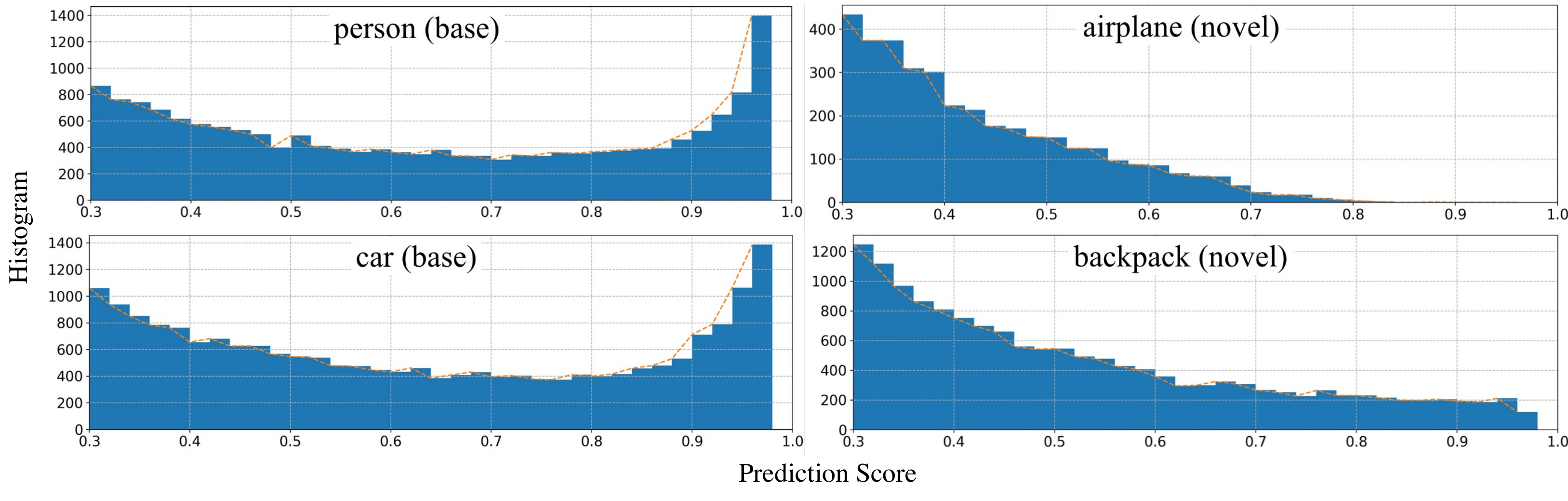}
         \caption{}
         \label{fig:confidence_biase}
    \end{subfigure}
    \vspace{-1.0em}
    \caption{
    (a): Pre-trained CLIP model~\cite{clip} produces inaccurate proposal-concept pairs in the left image, which are well refined by our Online Proposal Mining~(OPM) in the right image.
    (b): Under confidence of novel category predictions is observed from the histograms of confidence scores on COCO~\cite{coco_zeroshot}.
    %%%%%%%%%%%% cook %%%%%%%%%%%%%%%%
    % 这里（a）可以把caption加上，毕竟intro里也有用caption做说明；
    % （b）横纵坐标离坐标系太远了，容易看不懂，是否可以考虑少加点儿表，把横纵坐标写全，多余的表放补充材料？
    %  （b）的另外一个方案，所有concept都画在一个表上，不同concept用不同颜色表示
    % 字体太小了
    %%%%%%%%%%%%%%%%%%%%%%%%%%%%%%%%%%
    }
    \vspace{-1.5em}
    \label{fig:teaser}
\end{figure}

%Typical OVD methods~\cite{vild,regionCLIP,detic} first learn an unbounded vocabulary of concepts from image-caption pairs, and then transfer the general vision-language knowledge to facilitate OVD with detection annotations of base categories alone. 
Typically, OVD methods~\cite{ovrcnn, vild,regionCLIP,detic} are accomplished by first excavating an unbounded vocabulary of concepts as well as their general vision representations from image-caption pairs, and then transferring the general vision-language knowledge 
%{\color{red}facilitate OVD typically featured with detection annotations of base categories alone.}
% facilitate a detector. Typically, the detector has detection annotations of base categories alone.
%
to the common detection learning process on the box-annotated data with base categories alone.
Albeit their promising progress, we realize that two critical facts are sorely neglected in most existing methods including \textit{proposal-level vision-language alignment and base-novel category prediction equalization}, 
%{\color{red}Consequently, the performance increase of OVD has remained stagnant so far.}
which hamper the performance boost of existing OVD methods.

%In terms of proposal-level vision-language alignment, 
For the first issue, it is widely known that proposal annotations are the most beneficial hints for object detection. 
However, off-the-shelf studies dwell on weaker image-level vision-language annotations since most VLMs~\cite{clip,pixelbert,gao2022pyramidclip} are pre-trained upon a whole caption text describing an entire image.
Inaccurate proposal-concept pairs stem from a direct extension to excavate proposal-level vision-language annotations.
Left of Fig.\,\ref{fig:proposal_problem} illustrate an experiment \emph{w.r.t}. CLIP~\cite{clip} where two inaccurate proposal-concept pairs arise: (1) A big ambiguity stems from the matching between located proposals and concepts of ``person'' and ``skis''. (2) The proposal of ``person'' only encloses part of the person within the image. 
Therefore, simply transferring the coarse vision-language knowledge to OVD models~\cite{ovrcnn} or exploiting the noisy proposal-concept pairs for fine-grained detection model, even with distillation~\cite{vild}, may interfere with accurate object localization and degrade proposal-level vision-language alignment. 
We realize that opulent concepts describing proposal information are widely distributed in image-level caption texts. For example, Fig.\,\ref{fig:proposal_problem} is in line with the caption ``a person jumping a pair of skis in the air'' where ``person'' and ``skis'' are indeed fine-grained proposal concepts. 
Exploiting these proposal concepts might help overcome inaccurate proposal-concept pairs and uphold accurate object localization.
Recent Object Centric OVD~\cite{objcentric-ovd} solves this issue by training on filtered proposal-concept pairs. %.and obtains 36.5 $AP_{50}$ upon novel categories of COCO~\cite{mscoco}. 
%
%Nevertheless, their performance gains are built upon an off-the-shelf detector MViT~\cite{Maaz2022Multimodal} which is pre-trained on datasets with large scale box annotations.
%, which violates the basic principle of OVD task. 
%
However, the performance gains rely on a pre-trained MViT detector~\cite{Maaz2022Multimodal} upon large-scale box annotations.

%due mostly to the awkwardness in excavating accurate vision \& language alignment in proposal level.

% 两方面不足介绍：
% ----- px 05/18
%Despite their promising progress, existing OVD methods have two main problems. \emph{First, % obtaining 
%acquiring proposal-level vision-language knowledge suitable for object detection is challenging.} Although pre-trained VLMs~\cite{clip,pixelbert,gao2022pyramidclip} have covered a wide range of concepts, the learned knowledge is restricted to the image level, which is insufficient for object detection. Furthermore, since the text representations of each category are learned implicitly from image captions, they typically yield suboptimal decision boundaries for OVD tasks. As shown in the left image of Fig.~\ref{fig:proposal_problem}, 
% CLIP~\cite{clip} selects highly overlapped proposals for ``person'' and ``skis'' due to their high similarity scores. 
%CLIP~\cite{clip} has large ambiguity to match correct proposals to ``person'' and ``skis'' concepts, 
%and the proposal of ``person'' only encloses part of the person. Therefore, simply transferring the coarse vision-language knowledge to OVD models~\cite{ovrcnn} or exploiting the noisy proposal-concept pairs for fine-grained detection model with distillation~\cite{vild} may interfere with accurate object localization and degrade the learning process of proposal-level vision-language alignment.

As for the second problem, we empirically observe that
%in terms of base-novel category prediction equalization, 
existing optimized OVD models are prone to making imprecise predictions biased towards base categories. An illustrative example is given in Fig.\,\ref{fig:confidence_biase} where we count instances \emph{w.r.t}. the prediction score from Detic model~\cite{detic} on COCO dataset~\cite{coco_zeroshot}.
As we can observe that Detic model can well distinguish base categories such as ``person'' and ``car'' with most instances being given a high prediction score. On the contrary, novel categories like ``airplane'' and ``backpack'' are often endowed with a very low prediction score which indicates a large amount of mispredicted instances. Moreover, we empirically find that most mispredicted novel instances are grouped into base categories. Thus, the performance of OVD is still far from satisfactory for real-world applications.
Basically, we attribute the prediction bias issue to OVD training paradigm and class-imbalanced training datasets. 
For the training paradigm, as pointed out by previous works~\cite{ovrcnn,regionCLIP}, the detectors in OVD are fine-tuned upon base categories, which often leads to catastrophic forgetting of the general vision-language knowledge in pre-trained VLMs, in particular to novel categories.
%{\color{orange}in particular to very small number of base categories.}
%
As for class imbalance, OVD task requires abundant base training instances collected beforehand.
Nonetheless, training samples for novel categories are scarcely excavated. Then, it causes insufficient learning of proposal-level vision-language knowledge for novel categories and the optimized detectors are biased towards base categories in inference.

\begin{figure*}
     \centering
     \includegraphics[width=0.9\linewidth]{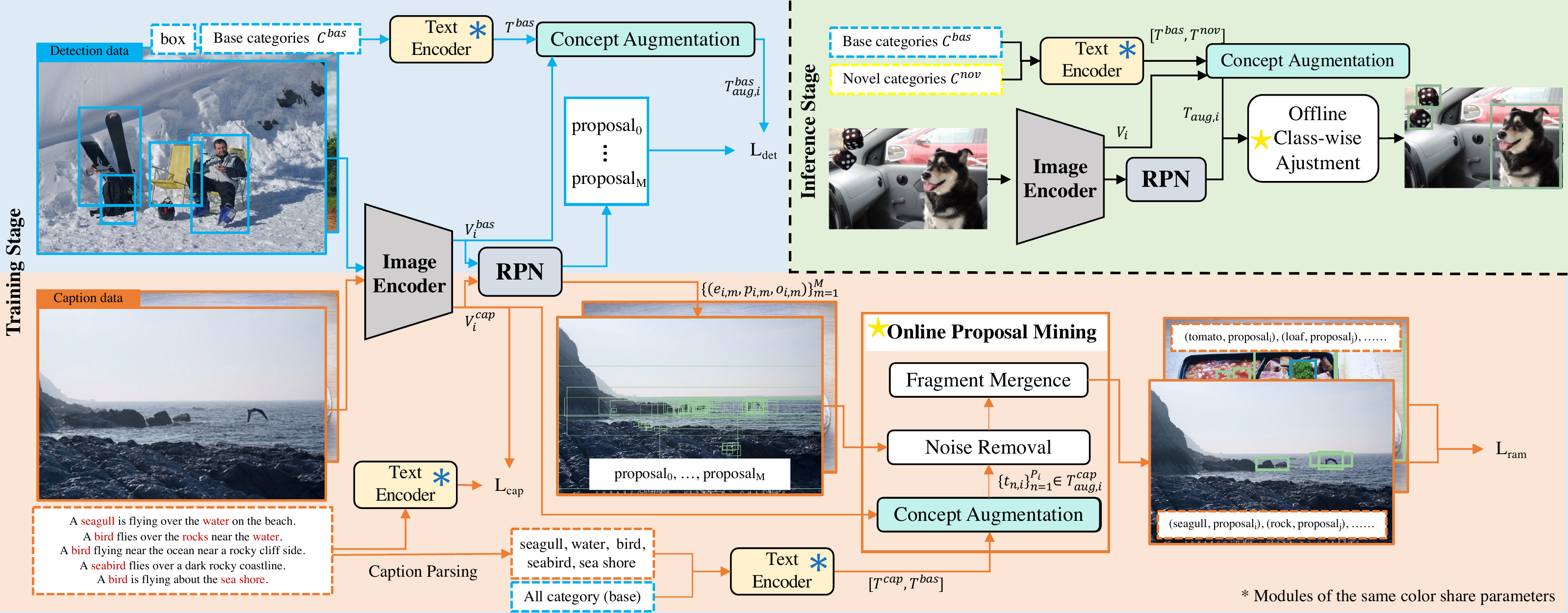}
     \vspace{-0.5em}
     \caption{The framework of our MEDet.
     With a mini-batch of data from both the detection dataset and the image-caption dataset, MEDet jointly trains object detection (\textcolor[RGB]{17,160,255}{blue} part) and learns rich proposal-level vision-language knowledge by conducting online proposal mining (\textcolor[RGB]{239,115,33}{orange} part) in an end-to-end manner. After the standard detection inference (\textcolor[RGB]{153, 204, 153}{green} part), we propose offline class-wise adjustment to handle the confidence bias between the base and novel categories. The \textbf{\textcolor[RGB]{10,112,192}{$\ast$}} means a frozen network. The \textbf{\textcolor[RGB]{255,232,0}{$\star$}} means contributions of this paper. }
    %\caption{The framework of our MEDet. Detection data with only base categories (\textcolor[RGB]{17,160,255}{blue} part) is used to train the detector. And caption dataset (\textcolor[RGB]{239,115,33}{orange} part) is used to learn rich %and complete 
    %proposals-level vision-language knowledge by conducting online proposal-level. During inference (\textcolor[RGB]{153, 204, 153}{green} part), we propose class-wise backdoor adjustment to handle the confidence bias. The \textbf{\textcolor[RGB]{10,112,192}{$\ast$}} means a frozen network. The \textbf{\textcolor[RGB]{255,232,0}{$\star$}} means contributions of this paper.}
     \label{fig:framework}
     \vspace{-1.0em}
\end{figure*}

In this paper, we present MEDet, a novel proposal \textbf{M}ining and prediction \textbf{E}qualization framework for open-vocabulary object \textbf{Det}ection, to overcome the above two issues. Fig.\,\ref{fig:framework} depicts the overall framework of our MEDet where proposal-level vision-language alignment and base-novel category prediction equalization are respectively accomplished via an online proposal mining (OPM) in the training and via an offline class-wise adjustment (OCA) in the inference.
The OPM filters out low-qualified alignments in a three-step coarse-to-fine fashion. 
%
%To eliminate the information variance between texts and images, it first augments text embedding discrimination by online interacting with the corresponding image embedding information via a cross-modality based transformer block\,\cite{ViT}. 
It first augments text embedding discrimination by online interacting with the corresponding image embedding information via a cross-modality-based transformer block\,\cite{ViT}. 
%
%Then, with the augmented concept embeddings, it further removes noisy proposal-concept pairs by building their semantic similarity distribution.
Then, it further removes noisy proposal-concept pairs by building their semantic similarity distribution.
%
%Finally, to clean proposal fragments holding broken objects, OPM further takes into consideration the spatial relationship between two proposals which are merged into one if they are of high overlapping.
Finally, to clean proposal fragments holding broken objects, OPM further merge two proposals if they are of high overlapping.
Built upon OPM, an Iterative Matching with Recurrent Attention Memory (IMRAM)\,\cite{imram}  is used to discover the full latent proposal-level vision-language alignments, resulting in more accurate proposal-concept pairs than common pre-trained VLMs, as shown in the right of Fig.\,\ref{fig:proposal_problem}.
%
%
%

% ----- px 05/18
%{\color{blue}%In this paper, we present MEDet (Proposal \textbf{M}ining and Prediction \textbf{E}qualization for Open Vocabulary Object \textbf{Det}ection), a novel framework to overcome the two problems for better OVD performance. The overall framework is illustrated in Fig.~\ref{fig:framework} in Sec.~\ref{sec:Method}.
%
%For the first problem, we conduct Online Proposals Mining (OPM) in the end-to-end OVD network at the training stage to fully explore proposal-level vision-language knowledge from the image-caption information. 
%Specifically, we first leverage a caption parser and an RPN to obtain noun concepts in the caption and object proposals in the image, respectively. For high-accuracy proposal mining, we first augment text embeddings of noun concepts by online feature interaction with the corresponding image features. 
%Then, noisy proposal-concept pairs are removed by analyzing their semantic similarity distribution and spatial relationship. 
%Finally, a recurrent attention memory-based contrastive learning module is used to discover the full latent proposal-level vision-language alignments. Overall, OPM enables more accurate proposal-concept pairs compared with common pre-trained VLMs, as demonstrated in the right image of Fig.~\ref{fig:proposal_problem}.

% peixianchen-20221102
% In addtion, to mitigate the under-confidence issue on novel classes, we devise a simple yet effective debiasing scheme. Due to lack of proposal-level annotations in caption dataset, 
% peixianchen-20221102

After training the OVD model, our OCA module further post-processes the trained OVD model for a better prediction. It performs density-based clustering~\cite{rodriguez2014clustering} upon all dataset proposals in compliance with the same predicted concept. And then a de-bias term vector is offline computed based on the clustering density and online deployed to adjust the prediction of an incoming proposal.
Experiments on COCO~\cite{coco_zeroshot} and LVIS~\cite{lvis} benchmarks demonstrate that our method outperforms other cutting-edge OVD methods~\cite{ovrcnn,vild,detic,regionCLIP}.
For example, MEDet reaches $32.6\%$ AP50 for novel categories on the COCO dataset, which considerably suppresses the \textit{state-of-the-art} approaches by $3.5\%$.

%{\color{orange}In addition, to mitigate the under-confidence issue on novel classes, we devise a simple yet effective debiasing scheme. 
%Motivatied by the method of post-hoc logit adjustment~\cite{menon2020logitadjustment}, we propose an offline class-wise adjustment (OCA) module to produce equalized predictions. 
%Specifically, we through offline calculate the density of all concepts of caption dataset to estimate the class-wise bias. At the inference stage, we leverage this bias to adjust the output for less biased prediction. 
%Specifically, we conduct density-based clustering for all concepts on the final predictions of the training caption dataset. Based on the density distribution of each concept, the class-wise bias is estimated conveniently. At the inference stage, we leverage this bias to adjust the output for less biased prediction. }

% By reconstructing the issue from the causal intervention~\cite{tang2020unbiased,yang2021deconfounded,yue2021counterfactual,niu2021counterfactual}, we design a class-wise backdoor adjustment (CBA) module to produce equalized predictions. %in the inference. 
% Specifically, at the inference stage, we estimate the class-specific confounder embedding via density clustering on the training dataset, and leverage it to adjust the output for less biased prediction. 

% 
%Our major contributions can be summarized as follows:
%\begin{itemize}
%\itemsep 0cm
%    \item We propose a novel online proposal mining method to fully exploit the caption information for learning proposal-level vision-language knowledge toward open vocabulary object detection.
    
%    \item We devise an effective offline adjustment scheme to alleviate the under-confidence issue and promote the detection results on novel classes. %, achieving better predictions.
    
%    \item Extensive experiments show that our proposed framework achieves new state-of-the-art performance on the novel categories when using the same caption dataset and base categories. 
%\end{itemize}

%}
%%%%%%%%%%%%%%%%%%%%%%%%%%%%%%%%%%%%%%%%%%%%%%%%%%%%%%%%%%%%

\section{Related work}

%\noindent\textbf{Open-Vocabulary Object Detection (OVD).}
\textbf{Open-Vocabulary Object Detection}.
%%%%davina
Typically, researchers scale up the vocabulary size for object detection by exploiting rich knowledge within pre-trained VLMs~\cite{clip,pixelbert,gao2022pyramidclip}.
OVR-CNN~\cite{ovrcnn} first learned a projection layer behind the backbone of Faster R-CNN~\cite{fasterrcnn} to align visual space with textual space~\cite{pixelbert}, and then fine-tuned the detector with only base categories.
Distillation-based approaches~\cite{vild,zsdyolo} aligned the vision extractors with both image and text encoders of CLIP~\cite{clip}. 
% Recently, Zhong~\etal\cite{regionCLIP} tackled the domain shift problem when adopting VLMs~\cite{clip,pixelbert} %, because the models are trained to match an image as a whole to a text description, without capturing the fine-grained alignment between image regions and text spans.
%To alleviate the problem, they
% and proposed RegionCLIP~\cite{regionCLIP}: it first leverages a CLIP model to match image regions with template captions, and pre-trains the model to align the region-concept pairs in the feature space, and finally transfers the pre-trained model to object detection.
RegionCLIP~\cite{regionCLIP} first leveraged CLIP model to match image regions with template captions, then pre-trained the model to align the region-concept pairs in the feature space and finally transferred the pre-trained model to downstream object detection.
However, its training pipeline is complex, and the pseudo region-concept pairs are noisy.
In this paper, we propose online proposal mining~(OPM) within an end-to-end OVD network and simplify the training pipeline.
Moreover, OPM provides more reliable region-concept pairs for acquiring rich proposal-level vision-language knowledge.
%\textcolor{gray}{In this paper, we propose dynamic proposal mining with concept augmentation to obtain more reliable region-text pairs, and discover the full latent alignments with a stacked cross attention based contrastive learning module.}
Recently, Zhou~\etal\cite{detic} applied additional
image-level supervision (\textit{e.g.}, predefined classification categories) to the largest proposal and did not supervise other outputs for image-labelled data. 
A limitation is that it requires image-level annotations within a predefined hand-crafted taxonomy and only learns the classes within the taxonomy. %In contrast, %following OVR-CNN,
%we leverage a larger vocabulary of concepts within captions that can generalize well to target classes.
% we use all concepts within captions that are more open-vocabulary, %and also more prevalent on the web,
% and we learn to generalize to any set of target classes on demand, without having to know them beforehand. 
%%%%%davina

%  classimbalance的处理方法以及Ovd解决训练bias的问题的方法（如ovrcnn freeze v2l，regionclip 修改基类的focalloss参数）。
%\noindent\textbf{Imbalance Debiasing Strategies in Object Detection.}
\textbf{%Prediction
Debiasing Strategies in Object Detection}.
Real-world detection datasets~\cite{coco_zeroshot,lvis,pan2020DRN} typically exhibit class-imbalance label distributions, %which make
making it difficult to learn generalized representation across classes.
% To improve overall results on both frequent classes and rare ones, researchers have proposed inspiring methods.
Existing debiasing approaches can be roughly categorized into four mainstreams: (1) re-weighting~\cite{chang2021resampling,lvis,shen2016relay} to sample more instances of rare classes; 
(2) 
cost-sensitive learning~\cite{tan2020equalizationloss,tan2021equalizationlossv2,wang2021seesawloss} to spare more attention on hard examples;
(3) knowledge transfer~\cite{jamal2020rethinking,kang2019decoupling,liu2019large} to learn generalized representation progressively;
and (4) post calibration~\cite{li2020balancedgroupsoftmax,menon2020logitadjustment,pan2021norcal,tang2020causal} to refine the model's output during the inference. %for balanced predictions.
% Basically,
To apply the above methods, we need the information on the rare target classes (\emph{e.g.}, class frequency) in advance.
% However, 
Unfortunately, 
in the setting of OVD~\cite{vild,ovrcnn}, we %don't know the
do not have access to the 
definition of \textit{novel} categories during training, let alone class frequencies. % or their class frequency during inference.
The scarcity of information makes it difficult and ineffective to leverage the popular imbalance debiasing schemes. %to boost open-vocabulary detectors.
Besides, the prediction bias also results from the training strategies of OVD models. Existing methods adopt model freezing~\cite{ovrcnn} and focal scaling~\cite{regionCLIP} to alleviate the 
catastrophic  
forgetting the general knowledge %concepts
in the pre-trained VLMs when fine-tuning the detectors on the base categories.
%{\color{red}In this paper, building on the theory of causal  intervention~\cite{pearl2000models,glymour2016causal,niu2021counterfactual}, we present a novel and effective backdoor adjustment scheme in the inference to eschew the bias for equalized predictions and enhance overall results on both base classes and novel classes.}
%{\color{orange}In this paper, building on the Bayes-optima theory for post calibration, we directly estimate the class-wise bias based on the density distribution of each concept in the caption dataset, instead of relying on the information of class frequencies.} 
%% region-clip 里面把 focal loss 用在了 transfer learning 阶段、是假设知道 哪些是 novel class；不在 OVD 的对比范围内
%\textcolor{gray}{Existing OVD works handle the issue with the aforementioned methods (\emph{e.g.}, model freezing~\cite{ovrcnn}) but the improvement is marginal.}
%In this paper, building on the theory of causal  intervention~\cite{pearl2000models,glymour2016causal,niu2021counterfactual,yang2021deconfounded}, we present a novel and effective backdoor adjustment scheme to eschew the imbalance bias for equalized predictions and enhanced overall results on both base classes and novel ones.

%%%%%%%%%%%%%%%%%%%%%%%%%%%%%%%%%%%%%%%%%%%%%%%%%%%%%%%%%%%%
\section{Methodology}
\label{sec:Method}

\subsection{Preliminary}
\label{sec:Preliminary}

Considering a base dataset for object detection $\mathcal{D}^\mathrm{bas} = \big\{ (I_i^\mathrm{bas}, \{(b^{k}_{i}, c^{k}_{i})\}_{k=1}^{K_i} ) | c^k_i \in C^\mathrm{bas} \big\}_{i = 1}^{N^\mathrm{bas}}$ where $I_i^\mathrm{bas}$ denotes the $i$-th image containing $K_i$ bounding box and class pairs $(b_{i}^{k}, c_{i}^{k})$, and $C^\mathrm{bas}$ denotes the base category set, OVD intends to learn a detector that recognizes well-unseen novel categories $C^\mathrm{nov}$ with $C^\mathrm{nov} \cap C^\mathrm{bas} = \phi$.
To this end, an image-caption dataset $\mathcal{D}^\mathrm{cap} = \{ (I_i^\mathrm{cap}, s_i, \{w^k_i\}_{k=1}^{P_i} ) | w^k_i \in C^\mathrm{cap} \}_{i=1}^{N^\mathrm{cap}}$ is introduced in VLMs-based OVD methods~\cite{ovrcnn, detic,regionCLIP} in which $s_i$ is a caption sentence describing the image $I^\mathrm{cap}_i$, $\{w^k_i\}_{k=1}^{P_i}$ is a set of word concepts (nouns) extracted from $s_i$, and $C^\mathrm{cap}$ is the union of all concepts in $\mathcal{D}^\mathrm{cap}$.
Then, the OVD models are trained upon both $\mathcal{D}^\mathrm{bas}$ and $\mathcal{D}^\mathrm{cap}$. Typically, the overall training loss in existing studies can be generally formulated as follows:
\begin{equation}
\begin{split}
    \mathcal{L}_\mathrm{ovd}
     = & \mathcal{L}_\mathrm{rpn}(I^\mathrm{bas}_i, \{b^k_i\}_{k=1}^{K_i}) + \mathcal{L}_\mathrm{cls}(I_i^\mathrm{bas}, \{c^k_i\}_{k=1}^{K_i}) \\
     + & \mathcal{L}_\mathrm{reg}(I^\mathrm{bas}_i, \{b^k_i\}_{k=1}^{K_i}) + \mathcal{L}_\mathrm{bce}(I^\mathrm{cap}_{i}, s_i),
\end{split}
\label{all}
\end{equation}
where $\mathcal{L}_\mathrm{rpn}$ denotes the constraints for RPN, $\mathcal{L}_\mathrm{cls}$ is the classification loss and $\mathcal{L}_\mathrm{reg}$ regularizes the bounding box regression. These first three constraints are enforced upon the base dataset $\mathcal{D}^\mathrm{bas}$. As for $\mathcal{L}_\mathrm{bce}$, it represents the binary-cross entropy (BCE) applied to model embedding similarity of image $I^\mathrm{cap}_i$ and its caption sentence $s_i$.
%

%Compared with the base dataset $\mathcal{D}^\mathrm{bas}$, learning from the image-caption dataset $\mathcal{D}^\mathrm{cap}$ is much tougher due mostly to the untouchable bounding box information. Off-the-shelf OVD models suffer from immoderate reliance on base dataset. Consequently, existing methods fail to model proposal-level vision-language alignment and base-novel category prediction equalization as analyzed in Sec.\,\ref{Sec:intro}.
%
%It is of great necessity to excavate more valuable insights from the image-caption dataset $\mathcal{D}^\mathrm\mathrm{cap}$ where each concept $t_{i, q}$ accords with $J_q$ proposals with $J_q \le k$.

%{\color{red}Note for OVD, the classifier's weights are replaced by the text embeddings of base class names, so that the novel class embeddings can be used during testing, without changing the semantic space~\cite{ovrcnn}.}

\subsection{Framework of MEDet}
\label{sec:LearningMEDet}

MEDet models proposal-level vision-language alignment in the training stage by an online proposal mining~(OPM) mechanism in Sec.\,\ref{sec:OPM} and achieves base-novel category prediction equalization in the testing stage via an offline class-wise adjustment~(OCA) in Sec.\,\ref{sec:CBA}.
%The aforementioned two goals are respectively accomplished in our training stage by proposing an online proposal mining (OPM) mechanism in Sec.\,\ref{sec:OPM} and in our testing stage by proposing an offline class-wise adjustment (OCA) Sec.\,\ref{sec:CBA}. 
Before diving into a detailed discussion, we first outline the framework of our MEDet, as illustrated in Fig.\,\ref{fig:framework}.

In the training stage, for incoming mini-batch data: 
\textbf{First}, we feed to a text encoder the category concepts of the base dataset and word concepts from current image-caption mini-batch samples to obtain their text embeddings $T^\mathrm{bas}$ and $T^\mathrm{cap}$.
Similarly, we have image embeddings $V^\mathrm{bas}$ and $V^\mathrm{cap}$ from an image encoder. 
\textbf{Second}, for the text embeddings $T^\mathrm{bas}$ and $T^\mathrm{cap}$, we boost their discriminative ability by injecting image information in Sec.\,\ref{sec:OPM}, results of which are denoted as $T_\mathrm{aug}^\mathrm{bas}$ and $T_\mathrm{aug}^\mathrm{cap}$. 
\textbf{Third}, the base data ($T^\mathrm{bas}_\mathrm{aug}$, $V^\mathrm{bas}$) are used for a standard two-stage detection training~\cite{fasterrcnn}.
Considering the inaccessibility of bounding boxes, the image-caption pair ($T^\mathrm{cap}_\mathrm{aug}$, $V^\mathrm{cap}$) further goes through a series of noise removal and fragment mergence in Sec.\,\ref{sec:OPM} 
%to derive a set of refined concept-proposal pairs $\{(t_{i,q}, \{e_{i,j}\}_{j=1}^{J_q})\}_{q=1}^{Q_{i}}$ for image $I^\mathrm{cap}_i$. Here, each augmented text embedding $t_{i,q} \in T_\mathrm{aug}^\mathrm{cap}$ corresponds with $J_q$ proposal embeddings $\{e_{i,j}\}_{j=1}^{J_q}$. 
%
to derive a concept set $\mathcal{T}^0_i = \{t_{i,q}\}_{q=1}^{Q_i}$ and a proposal set $\mathcal{E}^0_i = \{e_{i,j} \}_{j=1}^{\sum_{q=1}^{Q_i}J_q}$ for the $i$-th image $I^\mathrm{cap}_i$. Here, the $q$-th concept $t_{i,q} \in T_\mathrm{aug}^\mathrm{cap}$ pairs with $J_q$ proposal embeddings $\{e_{i,j}\}_{j = 1 + J_0 + J_1 + ... + J_{q-1}}^{J_0 + J_1 + ... + J_{q}}$ where $J_0 = 0$.
%
%
%
%For the $i$-th sample in image-caption dataset $\mathcal{D}^\mathrm{cap}$, we denote its concept set $\mathcal{T}^0_i = \{t_{i,q}\}_{q=1}^{Q_i}$ and proposal set $\mathcal{E}^0_i = \{\{ e_{i,j} \}_{j=1}^{J_q}\}_{q=1}^{Q_i}$.
Also, we let $\mathcal{T}^0_{/i}$ and $\mathcal{E}^0_{/i}$ denote all concept sets and proposal sets in $\mathcal{D}^\mathrm{cap}$ except these from the $i$-th samples.
With this proposal-level vision-language knowledge, the most common way to align these pairs is through one-by-one contrastive loss~\cite{regionCLIP}, which however suffers from performance damage if mismatched pairs stem from the knowledge set. To avoid this obstacle, we adopt the Iterative Matching with Recurrent Attention Memory (IMRAM)~\cite{imram} to mitigate the negative impact. 
IMRAM considers two independent Recurrent Attention Memory (RAM) blocks, \ie, $\mathrm{RAM}_{e}$ and $\mathrm{RAM}_{t}$, to augment $\mathcal{T}_i^0$ and $\mathcal{E}_i^0$ in an iterative manner. The $k$-th step is briefly described as:
\begin{equation}
\begin{aligned}
&e_{i,j}^{k}, \mathcal{E}^{k}_i = \mathrm{RAM}_{e}(\mathcal{E}^{k-1}_i, \mathcal{T}^0_i);\, \\
&t_{i,q}^{k}, \mathcal{T}^{k}_i = \mathrm{RAM}_{t}(\mathcal{T}^{k-1}, \mathcal{E}^0_i),
\end{aligned}
\end{equation}
where $e_{i,j}^{k}$ is a reconstruction of $e_{i,j} \in \mathcal{E}_i^0$ established upon the concept set $\mathcal{T}_i^0$ and similar for $t_{i,q}^k$. The cross-modal cosine similarity is computed as:
\begin{equation}
\begin{aligned}
%&F_{k}(e_{i,j}, T) = cos(e_{i,j}, c_{i,j}^{v,k}); \;
%F_{k}(E, t_{i,q}) = cos(c_{i,q}^{t, k}, t_{i,q}),\\
S^{k}(\mathcal{E}^0_i, \mathcal{T}^0_i) = \mathbb{E}_j(\cos(e_{i,j},\, e_{i,j}^k)) 
 + \mathbb{E}_q(\cos(t_{i,q}^{k},\, t_{i,q})).
\end{aligned}
\end{equation}
%
%\sum_{q=1}^{Q_{i}}J_{q}
%

Here, $e_{i,j}^k$ is built upon the concept set $\mathcal{T}_i^0$, therefore, the similarity is modeled between the proposal embedding $e_{i,j}$ and all the concepts, leading to a major difference to the one-by-one contrastive loss~\cite{regionCLIP}, details of which can be referred to~\cite{imram}. 
The final similarity between the set pair $\mathcal{E}^0$ and $\mathcal{T}^0$ is: $S(\mathcal{E}^0_i, \mathcal{T}^0_i) = \sum_k S^k$.
%
%\begin{equation}
%  S(\mathcal{E}^0_i, \mathcal{T}^0_i) = \sum_k S^k. 
%\end{equation}
%
%
%
Similarly, we obtain the similarity of $S(\mathcal{E}^0_i, \mathcal{T}^0_{/i})$ and $S(\mathcal{E}^0_{/i}, \mathcal{T}_i^0)$, both of which play as negative factors to raise up positive factor of $(\mathcal{E}^0_i, \mathcal{T}^0_i)$ in a contrastive style:
%
%
%Then, we take $T$ and $E$ from the same image to form the positive set pair, and regards the feature sets $T = \{t_{i,q}\}, E_{/i} = \{ e_{/i, j} \}$ from different images in a batch as the negative set pairs. The hard negative pair that has the largest similarity value among negative pairs is denoted by $*$. Then, the final loss is:
\begin{equation}
\begin{aligned}
\mathcal{L}_\mathrm{ram} = \sum_{i=1}^B[\delta -S(\mathcal{E}_i^0, \mathcal{T}_i^0) + S(\mathcal{E}_i^0, \mathcal{T}_{/i})]_{+} \\
 + \sum_{i=1}^B[\delta - S(\mathcal{E}_i^0, \mathcal{T}_i^0) + S(\mathcal{E}^0_{/i}, \mathcal{T}_i^0)]_{+},
\label{equ:imram}
\end{aligned}
\end{equation}
where $[x]_{+} = \max(x, 0)$, $\delta$ is the margin, and $B$ is the training batch size.
Combining Eq.\,(\ref{all}) and Eq.\,(\ref{equ:imram}) leads to our final training objective:
\begin{equation}
    \mathcal{L} = \mathcal{L}_\mathrm{ovd} + \mathcal{L}_\mathrm{ram}.
\end{equation}

As for the inference stage, we devise an offline class-wise adjustment in Sec.\,\ref{sec:CBA} to accomplish the goal of base-novel category prediction equalization.

\subsection{Online Proposal Mining}
\label{sec:OPM}
%
%We propose OPM to progressively learn the rich and fine-grained proposal-level vision-language knowledge within $\mathcal{D}^\mathrm{cap}$. As illustrated in Fig.~\ref{fig:framework}, it contains three steps: (a) concept augmentation for classification discrimination; (b) noisy proposal-concept pair removal by analyzing the semantic similarity distribution; (c) fragment proposals merging based on the spatial distribution.

In this subsection, we formally introduce our online proposal mining (OPM) that extracts fine-grained proposal-level vision-language knowledge from the image-caption dataset $\mathcal{D}^\mathrm{cap}$ to promote detection performance.
Our OPM is a three-step pipeline including concept augmentation, noisy removal and fragment mergence.

\textbf{Concept Augmentation}. Current OVD implementations~\cite{regionCLIP,vild,detic} train detectors by embedding a concept as the results of a prompt template, \emph{e.g.}, \textit{\underline{It is a photo of} [concept]}. The shared prompt templates lead to less discriminative text features since images are much more diverse~\cite{he2021masked} compared to highly-semantic and information-dense texts. 
%The shared prompt templates lead to less discriminative image features since images are much more diverse~\cite{he2021masked} comparing to highly-semantic and information-dense texts.
Though prompt ensemble~\cite{vild} or prompt design~\cite{DetPro} might be an alternative, they cannot eliminate the information variance between texts and images.
To mitigate this issue, at the end of the backbone, we append a cross-modality attention-based transformer block~\cite{ViT} to enhance the text embedding discrimination by injecting image embedding information. 
%
%Giving text embeddings $T = \{ T^\mathrm{bas}, T^{pos} \}$ and image embeddings $V = \{ V^\mathrm{bas}, V^{pos} \}$, the augmentation is formulated as:

Giving text embeddings $T = \{ T^\mathrm{bas}, T^\mathrm{cap} \}$  in a batch, the augmentation is dynamically adopted for each image $V_{i} \in V = \{ V^\mathrm{bas}, V^\mathrm{cap} \}$:
\begin{equation}
\begin{split}
       T'_{\mathrm{aug},i} = T + \mathrm{CA}(W^{q}T, W^{k}V_{i}, W^{v}V_{i}), \\
       [T^\mathrm{bas}_{\mathrm{aug}, i}, T^\mathrm{cap}_{\mathrm{aug}, i}] = T'_{\mathrm{aug}, i} + \mathrm{FFN}(T'_{\mathrm{aug}, i}), 
        % & CA(Q,K,O) = Softmax(\cfrac{Q \cdot K^T}{\sqrt{D}}) \cdot O, \\
\end{split}
\end{equation}
where $\mathrm{CA}$ denotes cross attention of different modalities; $\mathrm{FFN}$ is the feed-forward network consisting of two linear layers; $W^q$, $W^k$ and $W^v$ represent projection matrices. For a brief description, we omit the subscript $i$ for each image.
Given that the base image-category pair $(T_\mathrm{aug}^\mathrm{bas}, V^\mathrm{bas})$ are accomplished with bounding box information, they are directly used for a standard two-stage detection training as stated in Sec.\,\ref{sec:LearningMEDet}.
Differently, we further refine image-caption ($T^\mathrm{cap}_\mathrm{aug}$, $V^\mathrm{cap}$) by noise removal and fragment mergence due to their invisibility of bounding boxes.

\textbf{Noise Removal}.
Considering an image $I^\mathrm{cap}_i$, we have its augmented concept embedding set $\{t_{i, n}\}_{n=1}^{P_i} \in T^\mathrm{cap}_{\mathrm{aug},i}$ and image-level embedding $V^\mathrm{cap}_i \in V^\mathrm{cap}$. Then, as shown in Fig.\,\ref{fig:framework}, $V^\mathrm{cap}_i$ is sent to the RPN to derive a proposal set $\{(e_{i, m}, p_{i,m}, o_{i,m})\}_{m=1}^M$ where $e_{i,m}$ is the feature embedding of the $m$-th proposal, $p_{i, m} \in \mathbb{R}^4$ and $o_{i, m}$ are its coordinates and objectiveness score. To match a proposal $t_{i, n}$ with an accurate feature embedding $e_{i, m}$, one naive approach is to consider the cosine similarity. However, as shown in Fig.\,(\ref{fig:proposal_problem}), highly-overlapped proposals are often identified as different categories due to semantic confusion. Instead, we introduce a similarity entropy measurement.

We first measure the semantic similarity between $t_{i, n}$ and $e_{i,m}$ in relation to their cosine similarity and objectiveness score of $e_{i,m}$ as:
\begin{equation}
    % \small
    SC_{n, m} = \cos(t_{i,n}, e_{i,m}) \cdot o_{i, m}, 
    \label{equ:sc} 
\end{equation}
which is then normalized by the softmax function to obtain the similarity entropy for the proposal embedding $e_{i, m}$ as:
\begin{equation}
    % \small
    E_m = \mathrm{entropy}( \mathrm{softmax}(SC_{:,m})).
    \label{equ:SimilarityEntropy}
\end{equation}

Here, we would like to stress the efficacy of $E_m$. A large $E_m$ value manifests the proposal embedding $e_{i,m}$ is almost equally matched with every concept embedding, which in turn indicates poor proposal mining.
On the contrary, a small $E_m$ means $e_{i, m}$ is well matched with some particular concept, in which case $e_{i, m}$ should be preserved.

%To this end, we regard the entire image as a proposal and first discard these proposals whose similarity entropy is lower than that of the image proposal since an entire image is a very rough proposal therefore any poorer proposal can be safely removed. 
To this end, we regard the entire image as a proposal and first discard these proposals whose similarity entropy is larger than that of the image proposal since an entire image is a very rough proposal therefore any poorer proposal can be safely removed. Next, we match each concept embedding $t_{i, n}$ with proposals of the top-$3$ largest semantic similarity to form positive proposal-level vision-language pairs. 
Lastly, we further filter out pairs for each concept whose largest semantic similarity is smaller than that between the concept and the image proposal.
%
%Lastly, we further filter out pairs with unreliable concepts {\color{orange}whose largest semantic similarity is smaller than that between the image proposal and its $k$-th similar concept.}
%
Note that, we do not calculate the similarity entropy for each concept like that for each proposal in Eq.\,(\ref{equ:SimilarityEntropy}) to delete  unreliable proposal-concept pairs, mostly because each concept often matches several different proposals in the OVD task. Therefore, the value of similarity entropy fails to reflect the quality of concepts.

%consider the similarity entropy of each concept for deleting unreliable proposal-concept pairs due mostly to the fact that each concept often matches with several different proposals in OVD task. Therefore, the value of similarity entropy fails to reflect the quality of concepts.}

\textbf{Fragment Mergence}.
Albeit the removal of noisy proposal-concept pairs, it remains some fragments that merely contain part of an object. Therefore, we propose to merge these fragments in compliance with their spatial relationships. For each concept $t_{i,n}$, we compute the IoU$_{j,t}$ (intersection over union) between its two matched proposals $p_j$ and $p_t$ as the spatial similarity evaluation.
These two proposals are retained if IoU$_{j,t}$ is below a threshold $\theta^\mathrm{iou}$ and merged into a larger one otherwise. We repeat these steps until no proposals can be merged, 
obtaining the final concept set $\mathcal{T}^0_i = \{t_{i,q}\}_{q=1}^{Q_i}$ and proposal set $\mathcal{E}^0_i = \{e_{i,j} \}_{j=1}^{\sum_{q=1}^{Q_i}J_q}$ for the $i$-th image $I^\mathrm{cap}_i$ as utilized in Sec.\,\ref{sec:LearningMEDet}.
%Here, the $q$-th concept $t_{i,q} \in T_\mathrm{aug}^\mathrm{cap}$ pairs with $J_q$ proposal embeddings $\{e_{i,j}\}_{j = 1 + J_0 + J_1 + ... + J_{q-1}}^{J_0 + J_1 + ... + J_{q}}$ where $J_0 = 0$. 

%\noindent\textbf{Baseline.}
%For the detection data $\mathcal{D}^{det}$ (base categories only), we apply it in a traditional manner $\mathbf{L}_{det} = \mathbf{L}_\mathrm{rpn} + \mathbf{L}_\mathrm{cls} + \mathbf{L}_\mathrm{reg}$.
%$\mathbf{L}_\mathrm{rpn}$ denotes the constraints for RPN.
%After obtaining the features of all proposals, the detector classifies them into base categories with $\mathbf{L}_\mathrm{cls}$ and constrains them into ground truth bounding boxes with the $\mathbf{L}_\mathrm{reg}$.
% 
%For the caption data $\mathcal{D}^\mathrm{cap}$, we apply the binary cross-entropy ($BCE$) loss to learn a coarse vision-language space in the baseline.
%Given the embeddings of a caption $T^\mathrm{cap}$ and the image features $V^{img}$, we have $\mathbf{L}_\mathrm{cap} = \sum_{i} BCE(V^{img}_i \cdot T^\mathrm{cap}_i, i)$.

% Class-wise Backdoor Adjustment
% Backdoor Equalized Prompt
\subsection{Offline Class-wise Adjustment}
\label{sec:CBA}

Suppose a total of $Q$ concept embeddings are obtained upon the whole training set, denoted as $\mathcal{T}=\{t_q\}_{q=1}^Q$. For an incoming proposal embedding $e$ in inference, OVD performs a linear projection from visual features to concept embeddings and prediction scores of $e$ are obtained as:
\begin{equation}
    % c = \underset{q}{\mathrm{argmax}} \;\; \mathcal{T} \cdot e^T.
    P(\mathcal{T}|e) = \mathcal{T} \cdot e^T.
\end{equation}

Nevertheless, as analyzed in Sec.\,\ref{Sec:intro}, the training paradigm and class-imbalanced datasets cause immoderate reliance of OVD models on the base dataset and damage the performance. 
Such OVD settings as well as the scarcity of information (\emph{e.g.}, the class name and frequency) of $C^\mathrm{nov}$ to test in inference, hinder the deployment of debiasing methods~\cite{tan2020equalizationloss,chang2021resampling,li2020balancedgroupsoftmax} for a better training paradigm.
%
%
%

%Motivated by Menon~\etal~\cite{menon2020logitadjustment} {\color{blue}which analyzes the Bayes-optimal problem and they present 
%a method of Post-hoc logit adjustment that explicitly modify a categories prediction scores based on its class priors}, we estimate the prediction distribution of all concepts in caption data without any proposal-level annotations and calculate the bias of them in the current model. In the inference stage, we diminish the bias within prediction scores to enhance overall results on both base classes and novel classes without any additional computational overhead.

% [观察 & 动机]

% [具体做法]

%%% 基于 density 聚类的两个超参数

% Based on the assumption of OVD, the vocabulary of concepts extracted from $\mathcal{D}^\mathrm{cap}$ includes all base categories and possible novel categories. And motivated by Menon~\etal~\cite{menon2020logitadjustment} 
% % \cite{menon2020logitadjustment}通过推理Bayes-optimal发现，类别的先验分布与最终分类的得分成比例，因此按照一定比例扣除类别的先验分布，则可以调整模型得分的平衡误差。
% {\color{blue}which analyzes the xxxx and they present a method of Post-hoc logit adjustment}, we propose an offline class-wise adjustment (OCA) scheme, which can estimate the prediction distribution of all concepts in caption data without any proposal-level annotations and calculate the bias of them in the current model. In the inference stage, we diminish the bias within prediction scores to enhance overall results on both base classes and novel classes without any additional computational overhead.

Inspired by ~\cite{menon2020logitadjustment} where the Bayes-optimal problem is analyzed in a post-hoc adjustment manner, we propose an offline class-wise adjustment~(OCA) to post-process the trained OVD model for a better prediction. For easy understanding, we first give our refined concept prediction of $e$ as:
\begin{equation}\label{de-bias}
    % c = \underset{q}{\mathrm{argmax}} \;\; \mathcal{T} \cdot e^T - \gamma \cdot \beta,
    P(\mathcal{T}|e) = \mathcal{T} \cdot e^T - \gamma \cdot \beta,
\end{equation}
where $\beta = \{\beta_q\}_{q=1}^{Q}$ denotes our de-bias term and $\gamma$ is a scalar to control de-bias degree.

After training, for the $q$-th concept $t_{q}$ associated with its proposal embedding set $\{e_i^q\}_{i=1}^{N_q}$ extracted from the entire training set, we perform density-based clustering~\cite{rodriguez2014clustering} which automatically forms $K_q$ cluster centers upon these proposals $\{e_i^q\}_{i=1}^{N_q}$, leading to an average of $\rho_q = \widetilde{N}_q/K_q$ proposal density for each cluster where $\widetilde{N}_q \le N_q$ since some outlier proposals will be removed by~\cite{rodriguez2014clustering}.

Then, the $q$-th de-bias term $\beta_q$ for concept $t_q$ is computed:
\begin{equation}
    \beta_q =\sqrt{K_q} \cdot \rho_q.
    \label{eq:PDB}
\end{equation}

We observe many common concepts such as ``dog'' and ``person'', embrace diverse images and result in a very large $K_q$ than others. The square root operation well prevents these illegitimate clustering. Our de-bias design is built upon a simple posterior truth:
a large number of proposals will be mistakenly attributed to concept $t_q$ to which we suppose the trained OVD model is prone, leading to either a large $K_q$ or $\rho_q$ and finally a stronger $\beta_q$ . Therefore, our design in Eq.\,(\ref{de-bias}) decreases the tendency of mistaken predictions.

Notice we do not conduct any optimization, thus our OCA is flexible for any novel categories. Besides, the estimation of de-bias term $\beta$ can be offline implemented once-for-all. Thus, it does not increase any computation burden in inference. Importantly, we find the calculated $\beta$ can be well reused on other OVD methods as listed in Tab.\,\ref{tab:CAug_CBA}.

\section{Experiments}
\label{sec:experiments}

\subsection{Setup}
\textbf{Datasets \& Metric}.
We evaluate our method on two standard open-vocabulary detection benchmarks modified from COCO~\cite{mscoco} and LVIS~\cite{lvis}. COCO Caption~\cite{cococaption} and Conceptual Caption (CC)~\cite{CC3M} are used respectively for OVD on COCO and LVIS to learn a large vocabulary of concepts $\mathcal{D}^\mathrm{cap}$.
COCO Caption has the same images and train/test split as the COCO Object dataset, which has $118,287$ images and $5\times$ captions.
We parse the captions by Scene-Graph-Parser~\cite{schuster2015scenegraph} and get $62,628$ noun concepts for COCO Caption and $76,311$ noun concepts for CC. 
On COCO, we follow the data split of~\cite{ovrcnn} with $48$ base categories $C^\mathrm{bas}$ and $17$ novel categories $C^\mathrm{nov}$, which are subsets of $80$ COCO object classes. The rest $15$ categories $C^\mathrm{ret}$ are also evaluated to further investigate the generalization of OVD models. On LVIS, following~\cite{vild}, we use the training/validation images and adopt the category split with $866$ base categories (common and frequent objects) and $337$ novel categories (rare objects). 
We adopt the standard object detection metrics: mean Average Precision (mAP) and AP50
. On COCO, we mimic the generalized setting~\cite{vscoco_zeroshot} and report AP50s for base and novel categories.
%On COCO, we report AP50 and use text embeddings of $C^\mathrm{bas}$ and $C^\mathrm{nov}$ as classifier weights to mimic the generalized setting~\cite{vscoco_zeroshot}.
On LVIS, we use a standard class-agnostic mask head~\cite{MaskRCNN} to produce segmentation masks for boxes. Following~\cite{detic}, the mask mAPs for novel categories and all categories are used for evaluation.

\begin{table}[t]
    \centering
    \caption{Results of OVD on COCO dataset~\cite{coco_zeroshot}.
    MEDet equipped with online proposal mining and offline class-wise adjustment outperforms other methods on the novel categories.}
    \resizebox{\linewidth}{!}{
    \begin{tabular}{l|cc|ccc}
        \toprule
        \multirow{2}{*}{Method} & \multicolumn{2}{c|}{Detector Training}  & \multicolumn{3}{c}{COCO Generalized (48+17)} \\
        &Backbone &Box generator  & Novel & \color{gray}{Base} & \color{gray}{All} \\
        \midrule
        Base-only (CLIP) & - & -  & 1.3 & \color{gray}{48.7} & \color{gray}{39.3} \\
        WSDDN~\cite{wsddn} & - & - & 20.5 &\color{gray}{23.4} & \color{gray}{24.6} \\
        Cap2Det~\cite{ye2019Cap2Det} & - & -  & 20.3 & \color{gray}{20.1}  & \color{gray}{20.1} \\
        PL~\cite{rahman2020PL} & RN50-FPN & COCO Base (48) & 4.12 & \color{gray}{35.9}  & \color{gray}{27.9} \\
        OVR-CNN~\cite{ovrcnn} & RN50-C4 & COCO Base (48) & 22.8 & \color{gray}{46.0}  & \color{gray}{39.9} \\
        HierKD~\cite{hierKD} & RN50-C4 & COCO Base (48) & 20.3 & \color{gray}{51.3}  & \color{gray}{43.2} \\
        ViLD~\cite{vild} & RN50-FPN & COCO Base (48) & 27.6  & \color{gray}{59.5} & \color{gray}{51.3}\\
        RegionCLIP~\cite{regionCLIP} & RN50-C4 & LVIS (1203) & 26.8 &  \color{gray}{54.8}  & \color{gray}{47.5}\\
        Detic~\cite{detic} & RN50-C4 & COCO Base (48)  & 29.1 &  \color{gray}{52.4} &\color{gray}{46.2}\\
        \midrule
        \rowcolor{gray!25} MEDet (Ours) &RN50-C4 & COCO Base (48) & \textbf{32.6} & \color{gray}{53.5} & \color{gray}{48.0} \\
        \bottomrule
    \end{tabular}
    }
    \label{tab:on_COCO}
\end{table}

\textbf{Implementation Details}.
\label{sec:implementation_details}
We leverage the CLIP text-encoder~\cite{clip} on ViT-B-32~\cite{dosovitskiy2020vit} to convert concepts to text embeddings. 
For COCO, we use Faster R-CNN~\cite{fasterrcnn} with the RN50-C4 configuration %to implement our system
and train $40,000$ iterations of batch size $4$ for the detection data and batch size $16$ for the caption data on 8 V100 GPUs. The learning rate is initially $0.02$ and multiplied by $0.1$ at the $25,000$ and $35,000$ steps.
% 具体训练参数
For LVIS, %following~\cite{detic},
we use CenterNet2~\cite{centernet} with the RN50-FPN architecture and the same data augmentation and learning schedules. 
All models are pre-trained for $30,000$ iterations on the detection dataset %(only base categories) 
and the caption dataset using loss $\mathcal{L}_\mathrm{all}$ in Eq.\,(\ref{all}).

%only detection loss $\mathcal{L}_{det}$ and caption loss $\mathcal{L}_{cap}$. 
% 
% In OPM, we choose the image-size proposal as a benchmark. An image-size proposal contain all objects, therefore we set the $\theta^{se}$ as the similarity entropy of the image-size proposal. The same, similarity between the image-size proposal and the $k$-th concept is set as the threshold $\theta^{sc}_k$.
To merge fragmented proposals, we set $\theta^\mathrm{iou}$ to $0.6$. And we set the margin $\sigma$ of $\mathcal{L}_\mathrm{ram}$ to $0.2$.
In OCA, we set $\gamma$ as $0.4$.
% For the empirical results of various $\gamma$ and the optimal one, please refer to Sec.~\ref{sec:ablation_investigation}.
For the detailed ablation results, please refer to Sec.\,\ref{sec:ablation_investigation}.

% px4 
\subsection{Main Results}
\label{sec:main_results}
\textbf{OVD on COCO}.
The comparisons of OVD results on the COCO dataset are shown in Tab.\,\ref{tab:on_COCO}.
``Base-only'' means training Faster R-CNN only on the detection data of base categories, and using text embeddings of class names from CLIP to replace the classifier's weights.
Compared with weakly supervised methods such as WSDDN~\cite{wsddn} and Cap2Det~\cite{ye2019Cap2Det}, and zero-shot methods PL~\cite{rahman2020PL}, our MEDet obtains a significant improvement on all metrics.
Compared with the OVD method OVR-CNN~\cite{ovrcnn}, our MEDet also demonstrates superiority (\emph{e.g.}, $32.6\%$ \vs $22.8\%$ on $C^\mathrm{nov}$).

\begin{table}[t]
   \centering
   \caption{Results of OVD on LVIS dataset~\cite{lvis}. MEDet better explores proposal-level vision-language knowledge on Conceptual Caption dataset and diminishes prediction bias among categories.}
    \footnotesize{
    \begin{tabular}{l|ccc}
        \toprule
        % \multirow{2}{*}{Method} & \multicolumn{2}{c}{LVIS} \\
        % & Novel & \color{gray}{All} \\
        Method & Distill & Novel & \color{gray}{All} \\
        \midrule
        WSDDN~\cite{wsddn} & $\times$ & 16.5  &\color{gray}{30.0}  \\
        ViLD~\cite{vild}   & $\surd$ & 16.8  &\color{gray}{25.2}  \\
        RegionCLIP~\cite{regionCLIP} & $\surd$ & 17.1  &\color{gray}{28.2}  \\
        DetPro~\cite{DetPro} & $\surd$ & 19.8 &\color{gray}{25.9} \\
        Detic~\cite{detic} & $\times$ & 21.0  &\color{gray}{30.9}  \\
        \midrule
        % \rowcolor{gray!25} MEDet (w/o CBA)  &21.3  &\color{gray}{31.0}  \\
         \rowcolor{gray!25} MEDet (Ours) & $\times$  & \textbf{22.4}  &\color{gray}{34.4} \\
        \bottomrule
    \end{tabular}
    \label{tab:on_LVIS}
    }
\end{table}

\begin{table}[t]
    \centering
    \caption{
    A comparison of generalization capability of OVD methods on retained 15 categories of COCO dataset~\cite{mscoco}.}
    \footnotesize{
    \begin{tabular}{l|ccc}
        \toprule
        Method & Retain & Novel & \color{gray}{All} \\
        %  & (15) & (17) & (80) \\
        \midrule
        OVR-CNN~\cite{ovrcnn}  & 11.5  & 22.9  & \color{gray}{38.1}  \\
        Detic-80~\cite{detic} & 11.5 & 27.3  & \color{gray}{38.3}  \\
        Detic-65~\cite{detic}  & 5.2 & 9.2  & \color{gray}{34.1} \\
        \midrule
        \rowcolor{gray!25} MEDet (Ours) & \textbf{18.6} & \textbf{32.6} & \color{gray}{42.4} \\
        \bottomrule
    \end{tabular}
    }
    \label{tab:coco_other_15}
\end{table}

As for the performance on \emph{novel} categories $C^\mathrm{nov}$, the core in OVD task, even using a weaker configuration the proposed MEDet outperforms other methods. For example,
%\textcolor{orange}{(\emph{i.e.}, a common backbone RN50-C4 pre-trained on ImageNet-1k~\cite{deng2009imagenet} rather than the powerful image encoder of CLIP)}, the proposed MEDet outperforms other methods.
ViLD~\cite{vild} adopts advanced training strategies (\emph{e.g.}, model distillation, model ensemble, and data augmentation), yet MEDet is substantially better on $C^\mathrm{nov}$ ($32.6\%$ \vs $27.6\%$) and still competitive on $C^\mathrm{bas}$ with a weaker backbone (RN50-C4 \vs RN50-FPN) and a simple training scheme.
RegionCLIP~\cite{regionCLIP} uses a stronger box generator trained on the large box-supervision dataset LVIS for generating proposal-concept pairs, while our MEDet conducts online proposal mining merely on the COCO dataset and acquires better proposal-level vision-language alignment, achieving $5.8\%$ AP50 gain on $C^\mathrm{nov}$.
The results of Detic~\cite{detic} are reproduced based on the official implementation and it labels the caption data with all the 80 categories in COCO.
Contrastively, without knowing novel categories beforehand in the training, MEDet achieves higher results on all metrics (\emph{e.g.}, $32.6\%$ \vs $29.1\%$ on $C^\mathrm{nov}$).

%even with the annotated data of novel categories in the training, yet the proposed MEDet achieves higher results on all metrics (\emph{e.g.}, $32.6\%$ \vs $29.1\%$ on $C^\mathrm{nov}$).
% \textcolor{blue}{potential confusion?}.
%% 这里的 32.6% & 29.1% 是 MEDet 和 Detic 的最佳结果吧？目前版本的 Tab.1 里面没有这个数字，可能会引起 confusion

\textbf{OVD on LVIS}.
To further verify the competitiveness of our method, we conduct comparison experiments on the LVIS benchmark~\cite{lvis}, as shown in Tab.\,\ref{tab:on_LVIS}.
%As we observe that our MEDet obtains significant improvements on the $337$ novel categories $C^\mathrm{nov}$ ($22.4\%$ \vs $16.8\%$, $17.1\%$, and $19.8\%$) using only $1.1$M samples from the CC dataset, which is only half the amount of data used by other competitive approaches~\cite{regionCLIP,DetPro}. 
Compared to recent OVD methods~\cite{regionCLIP,vild} that learn general proposal-level vision-language knowledge via distillation from CLIP, our MEDet online explores such knowledge  on the Conceptual Caption dataset in a self-driving way and
obtains significant improvements on the $337$ novel categories $C^\mathrm{nov}$ ($22.4\%$ \vs $16.8\%$, $17.1\%$).
Besides, our method also outperforms Detic ($22.4\%$ \vs $21.0\%$) without requiring one hand-crafted taxonomy. The comparisons demonstrate the effectiveness and flexibility of the proposed MEDet in the OVD scenario.

\textbf{Generalization of MEDet}.
Besides the evaluation experiments on $C^\mathrm{bas}$ and $C^\mathrm{nov}$, we also analyze the performance on the retained $15$ classes $C^\mathrm{ret}$ in the COCO dataset~\cite{coco_zeroshot} to further investigate the generalization ability of OVD models.
Tab.\,\ref{tab:coco_other_15} lists the results.
Note that, when using COCO Caption, %to learn the vocabulary of concepts,
Detic~\cite{detic} requires a hand-crafted taxonomy to map each concept to one class in a set $C^{T}$.
``Detic-80'' uses $C^\mathrm{bas} \cup C^\mathrm{nov} \cup C^\mathrm{ret}$ ($80$ classes) as $C^{T}$ 
and works well on $C^\mathrm{nov}$. But when $C^{T}$ is $C^\mathrm{bas} \cup C^\mathrm{nov}$ (65 classes), ``Detic-65'' performs worse in novel categories.
In contrast, our MEDet works better on both $C^\mathrm{nov}$ and $C^\mathrm{ret}$. The results ensure that the proposed method utilizes the caption dataset more effectively for generalization toward the OVD setting. %MEDet is generalized to more categories.

\begin{figure*}[t]
   \centering
   \begin{subfigure}[t]{0.22\textwidth}
        \centering
        \includegraphics[width=\linewidth]{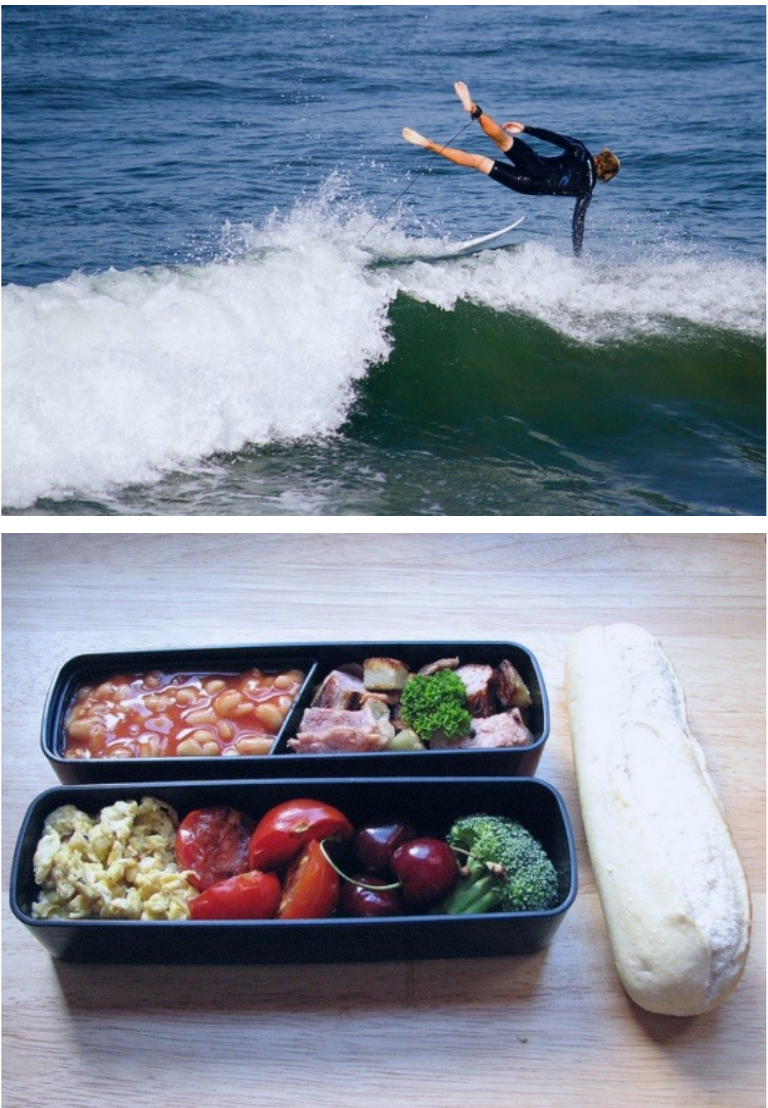}
        \caption{Original Images}
        \label{fig:proposal_opm_OriginalImage}
   \end{subfigure}
   \begin{subfigure}[t]{0.22\textwidth}
        \centering
        \includegraphics[width=\linewidth]{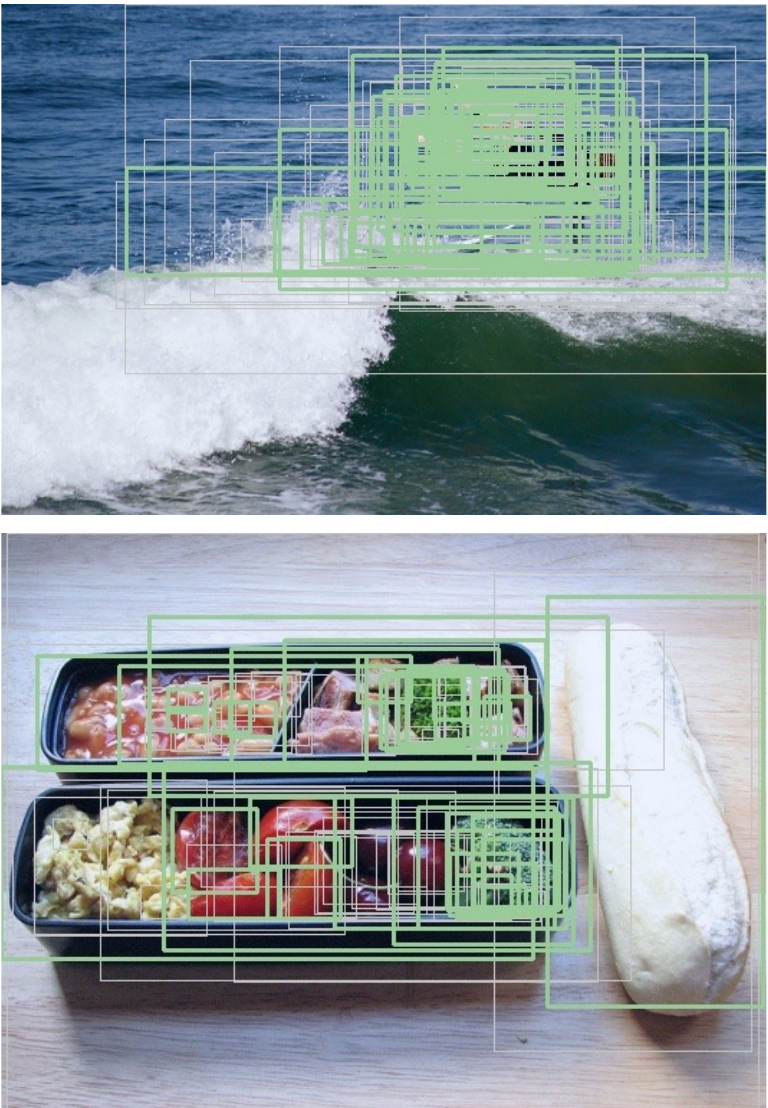}
        \caption{Proposal Filtering}
        \label{fig:proposal_opm_ProposalFilter}
   \end{subfigure}
   \begin{subfigure}[t]{0.22\textwidth}
        \centering
        \includegraphics[width=\linewidth]{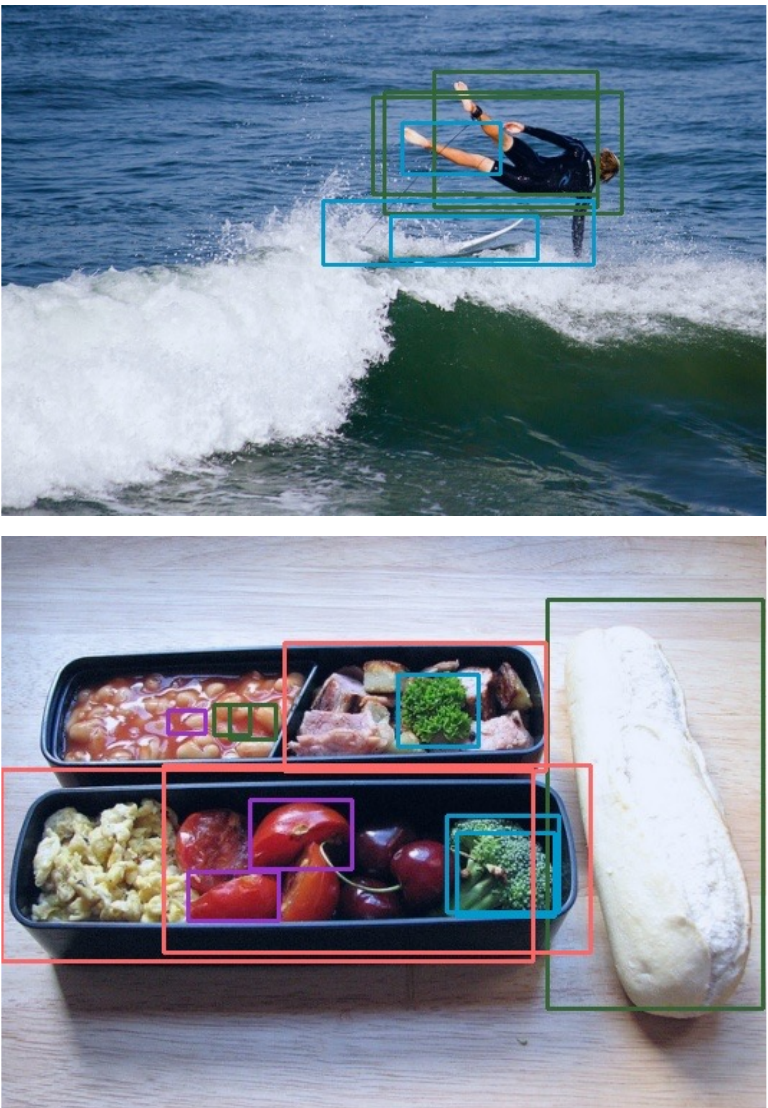}
        \caption{Top-3 Proposals}
        \label{fig:proposal_opm_top3}
   \end{subfigure}
   \begin{subfigure}[t]{0.22\textwidth}
        \centering
        \includegraphics[width=\linewidth]{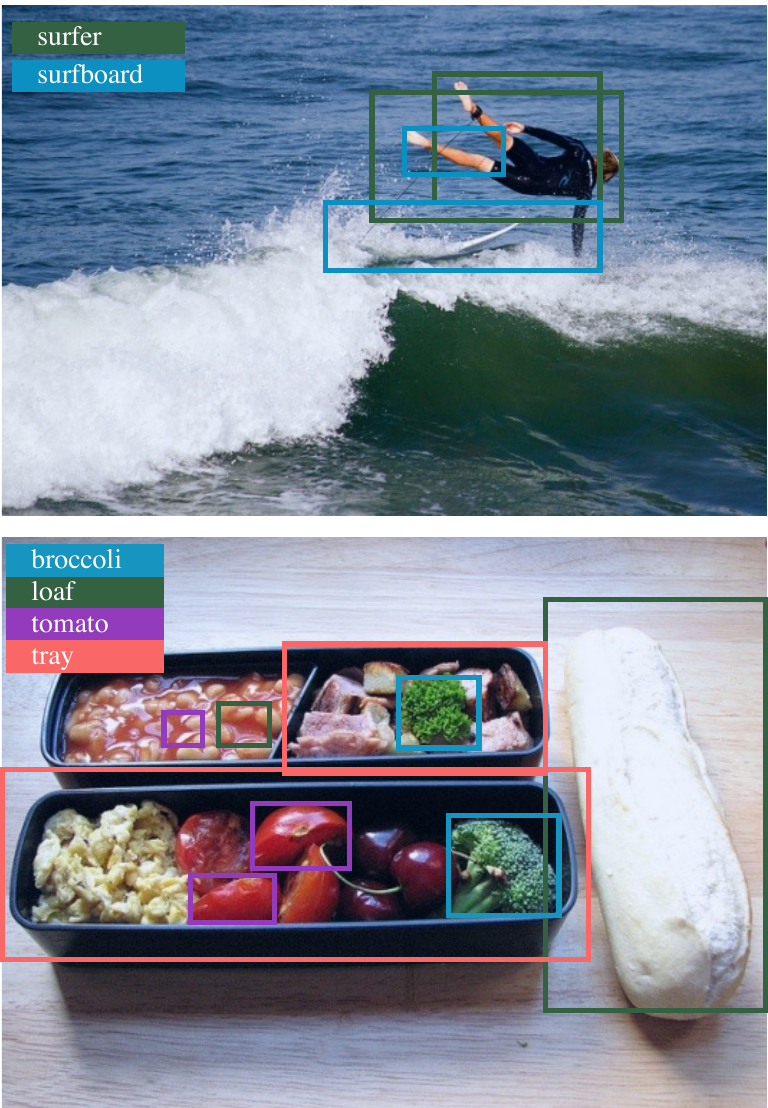}
        \caption{Proposal Merging}
        \label{fig:proposal_opm_ProposalMerge}
   \end{subfigure}
   \caption{Qualitative results of each step in OPM.
   (b) \textbf{Noise Removal}: Filter top-100 proposals from RPN by similarity entropy, where \textcolor[RGB]{153,204,153}{green} and \textcolor{gray}{gray} boxes are preserved and removed proposals, respectively.
   (c) \textbf{Noise Removal}: Select proposals with top-3 $SC$ scores for every concept.
   (d) \textbf{Fragment Mergence}: Merge proposals to eliminate fragmented proposals.}
   \label{fig:proposal_opm}
\end{figure*}

\subsection{Ablation Studies}
\label{sec:ablation_investigation}
\textbf{Components of MEDet}.
Tab.\,\ref{tab:ablation_on_MEDet} shows the ablation studies of components in MEDet. The pre-trained model only uses constraints $\mathcal{L}_{all}$, and is trained $30,000$ steps without Online Proposal Mining (OPM) and Offline Class-wise Adjustment (OCA).
% When we apply OPM (w/o CAug), the AP50 on novel categories reaches $31.0\%$ ($+13.4\%$). %(31.0\% \vs 17.6\%).
When we apply OPM, the AP50 on novel categories reaches $32.0\%$ ($+14.4\%$). 
The performance is also higher than other OVD methods in Tab.\,\ref{tab:on_COCO}. It means that OPM effectively explores proposal-level vision-language knowledge. 
% We add the Concept Augmentation (CAug) into OPM to effectively consider the semantic relationship among different concepts and the image embedding to adapt the text embeddings online. Then, the result on novel categories is increased by $1.0\%$. 
In the third row, the OCA handles the confidence bias %towards base classes
and thus boosts the performance on novel categories by $0.6\%$. % (32.6\% \vs 32.0\%).
Lastly, we replace the  matching loss~\cite{imram} (\emph{i.e.}, Eq.\,(\ref{equ:imram})) with a common grounding loss used by~\cite{ovrcnn} to verify the rationality of $\mathcal{L}_\mathrm{ram}$. Although the performance on novel categories drops by $1.8\%$ ($32.6\%$ \vs $30.8\%$), it is also competitive compared with the other methods in Tab.\,\ref{tab:on_COCO}.

\begin{table}[t]
    \centering
    \caption{Ablation studies of MEDet. Lines~2--3 show the increase in accuracy after the addition of our proposed OPM and OCA modules. For `MEDet w/o $\mathcal{L}_\mathrm{ram}$' we use a cross-modal attention used in OVR~\cite{ovrcnn} to learn the vision-language knowledge.}
    \footnotesize{
        \begin{tabular}{l|c|c|c}
            \toprule
            Method & Novel & \color{gray}{Base} & \color{gray}{All} \\
            \midrule
            Pre-trained model &17.6 &\color{gray}{43.3} &\color{gray}{36.6} \\
            % + OPM (w/o CAug) &31.0 &52.9 &\color{gray}{47.2}\\
            + OPM  & 32.0 &\color{gray}{53.1} &\color{gray}{47.5}  \\
            + OPM + OCA (MEDet)  &32.6 &\color{gray}{53.4} &\color{gray}{47.9}  \\
            \midrule
            MEDet w/o $\mathcal{L}_\mathrm{ram}$ &30.8 &\color{gray}{52.8} &\color{gray}{47.0} \\
            MEDet &32.6 &\color{gray}{53.4} &\color{gray}{47.9} \\
            \bottomrule
        \end{tabular}
        \label{tab:ablation_on_MEDet}
        }
\end{table}

\textbf{Effectiveness of OPM}.
In Tab.\,\ref{tab:ablation_on_OPM}, we remove the OCA module to verify the rationality of OPM. 
The result shows that Concept Augmentation can effectively consider the semantic relationship among different concepts and the image embedding to adapt the text embeddings online. When used, the result on novel categories is increased by $1.0\%$. 
When removing the Noise Removal step, we directly use proposals of the top-100 objectness scores to align with concepts. The result on novel categories is only $29.3\%$ ($-1.7\%$) due to noisy proposals. And disabling merging fragmented proposals may lose some large proposals, thus the performance decreases by $0.6\%$ ($30.4\%$ \vs $31.0\%$).

In order to further demonstrate the effectiveness of the OPM, we  show the qualitative results of each step in OPM.  Fig.\,\ref{fig:proposal_opm_ProposalFilter} shows the result after proposal filtering via similarity entropy( Eq.\,(\ref{equ:SimilarityEntropy})). The gray proposals are obtained according to the top-100 objectness scores from RPN. The green proposals are retained after removing incorrect proposals. 
Fig.\,\ref{fig:proposal_opm_top3} shows proposals of the TOP-3 $SC$ scores ( Eq.\,(\ref{equ:sc})) for every concept. Actually, we can select proposals for objects of different sizes and types. Fig.\,\ref{fig:proposal_opm_ProposalMerge} shows the refined proposals after fragment mergence. In a word, our OPM removes the most noise and obtains more reliable proposal-level vision-language knowledge.

\begin{table}[t]
        \centering
        \caption{Effectiveness of each step in OPM.}
        \footnotesize{
         \begin{tabular}{l|c|c|c}
            \toprule
            Strategy & Novel & \color{gray}{Base} & \color{gray}{All} \\
            \midrule
            w/o Concepts Augmentation & 31.0 &\color{gray}{52.9} &\color{gray}{47.2} \\
            w/o Noise Removal &29.3 &\color{gray}{52.4} &\color{gray}{46.5} \\
            w/o Fragment Mergence &30.4 &\color{gray}{52.8} &\color{gray}{47.0} \\
            \midrule
            OPM  &32.0 &\color{gray}{53.1} &\color{gray}{47.5}\\
            \bottomrule
        \end{tabular}
        }
        \label{tab:ablation_on_OPM}
\end{table}
    
\begin{table}[t]
    \centering
    \caption{Effectiveness of OCA introduced to other OVD methods.}
    \footnotesize{
    \begin{tabular}{l|c|c}
            \toprule
            Method & Novel & \color{gray}{All} \\
            \midrule
            OVR-CNN~\cite{ovrcnn} &22.8	&\color{gray}{44.3} \\
            % OVR-CNN~\cite{ovrcnn} + CAug &24.0	&\color{gray}{45.7} \\
            OVR-CNN~\cite{ovrcnn} + OCA &25.8	& \color{gray}{45.3} \\
            \midrule
            Detic~\cite{detic} &28.7	&\color{gray}{45.1} \\
            % Detic~\cite{detic} + CAug &29.6	&\color{gray}{46.6} \\
            Detic~\cite{detic} + OCA &29.8	&\color{gray}{46.8} \\ 
            \bottomrule
    \end{tabular}
    }
    \label{tab:CAug_CBA}
    \vspace{-1em}
    
\end{table}

\textbf{Effectiveness of OCA}.
Tab.\,\ref{tab:CAug_CBA} shows the improvement of OCA when it is used in different OVD models.
As the results of rows 2--4 (+ OCA), the improvements in both novel and base categories are generally higher than $1\%$. 
Notably, in the OCA experiment, we don't recalculate the bias according to different models. Instead, the bias $\vec{\beta}$  obtained from our MEDet model is reused directly to the inference stage of OVR-CNN and Detic, which fully demonstrates that the bias of concepts obtained from OCA can be reused by other OVD methods when training on the same caption dataset. 

\begin{figure}[tbp]
  \centering
  \begin{subfigure}[t]{.48\textwidth}
        \centering
        \includegraphics[width=\linewidth]{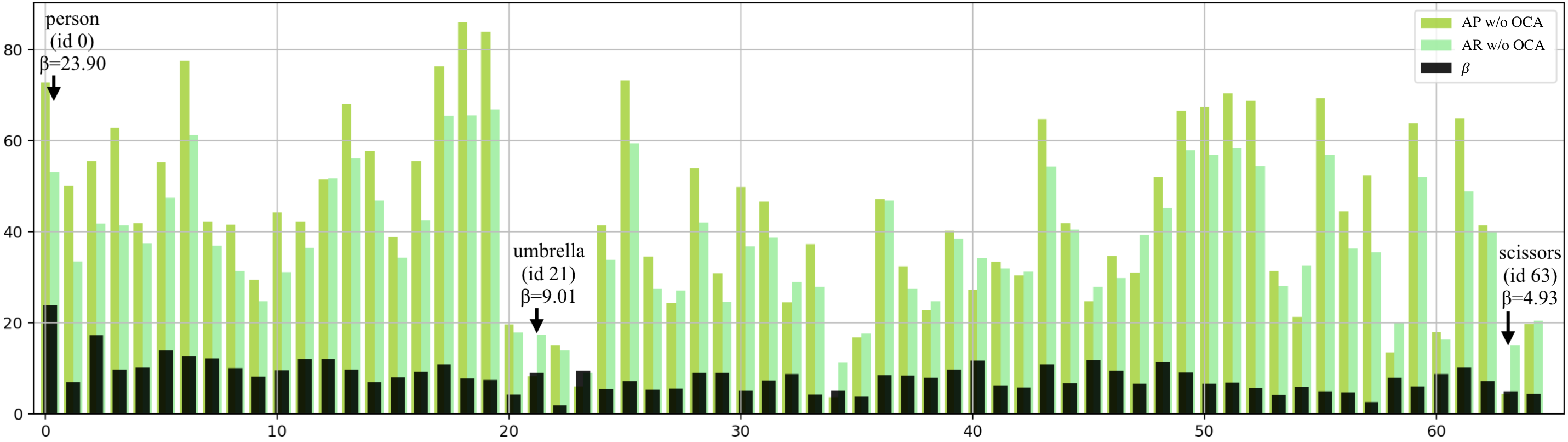}
        \caption{}
        \label{fig:AP_AR_woCBA}
  \end{subfigure}
  \begin{subfigure}[t]{.48\textwidth}
        \centering
        \includegraphics[width=\linewidth]{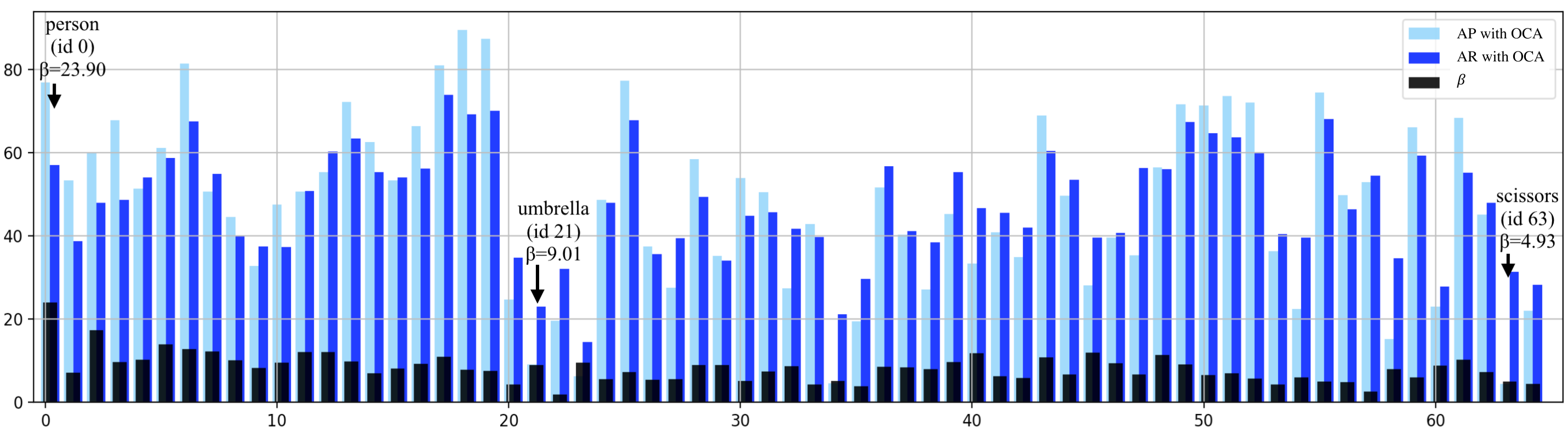}
        \caption{}
        \label{fig:AP_AR_with_CBA}
  \end{subfigure}
  \caption{\textbf{The AP, AR, and $\beta$ for each category on COCO~\cite{coco_zeroshot} when \textbf{(a)} we don't apply OCA or \textbf{(b)} we apply OCA.}
  The results indicate the positive contribution from OCA to the OVD task.}
  \label{fig:AR_AP_beta_CBA}
\end{figure}

To further understand the rationale and effectiveness of the proposed OCA, we analyze the AR, and AP, and estimate $\beta$ for each category. The results are shown in Fig.~\ref{fig:AR_AP_beta_CBA}.
As we observe that: for the category with lower $\beta$, such as umbrella (id 21, $9.0\%$ AP50, $18\%$ AR, $\beta=9.01$) and scissors (id 63, $4.5\%$ AP50, $16\%$ AR, $\beta=4.93$), the optimized model generates under-confident predictions, and generally has sub-optimal AP or AR performance, thus we should conduct less debiasing adjustment $\beta$ on it. After adjusting by OCA, the AP and AR on novel categories are improved, especially AR.
On the other hand, for the category with higher $\beta$, such as ``person'' (id 0, $76.8\%$ AP50, $53\%$ AR, $\beta=23.9$)), the model produces over-confidence outputs, and thus has higher AR performance but inferior AP performance.
So we need to conduct more debiasing adjustments on it.
After using OCA, the AP and AR on base categories are also improved. 
With the help of our OCA, we handle the confidence bias well and promote the overall OVD result.

\subsection{Hyper-parameter Analysis}

\textbf{Different $\theta^\mathrm{iou}$ in OPM}.
To investigate the $\theta^\mathrm{iou}$ in OPM, we analyze the AP50 results on COCO dataset with varied $\theta^\mathrm{iou}$ in Tab.~\ref{tab:ablation_on_IOU}. The observations show that a small threshold ($\theta^\mathrm{iou}=0.2$) causes too many proposals to be merged and the accuracy is reduced. And a high threshold ($\theta^\mathrm{iou}=0.8$) makes only a few (or no) proposals be merged.

% \noindent\textbf{The rationale and effectiveness of OCA.}
% To further understand the rationale and effectiveness of the proposed OCA, we analysis the AR, AP, and estimated $\beta$ for each class. The results are shown in Fig.~\ref{fig:AR_AP_beta_CBA}.
% As we observe that: for the class with lower $\beta$ (\emph{i.e.}, prior distribution), such as umbrella and scissors, the optimized model generates under-confident predictions, and generally has sub-optimal AP or AR performance, thus we should conduct less debiasing adjustment on it.
% On the other hand, for the class with higher $\beta$, such as person, the model produces over-confidence outputs, and thus has higher AR performance but inferior AP performance. So we need to conduct more debiasing adjustment on it.
% With the help of our OCA, we handle the confidence bias well and promote the overall OVD result.

\textbf{Different $\gamma$ in OCA}.
To investigate the optimal $\gamma$ of Eq.(\ref{eq:PDB}) in OCA, we analyze the OVD performance with varied $\gamma$. The AP50 results on COCO dataset~\cite{coco_zeroshot} are listed in Tab.\,\ref{tab:effect_of_gamma_in_CBA}.
The observations show the optimal $\gamma$ and indicate that the positive effect of OCA is relatively insensitive to $\gamma$.

\begin{table}[t]
    \centering
    \caption{The results of various $\theta^\mathrm{iou}$ in OPM on COCO.}
    \footnotesize{
    \begin{tabular}{l|ccccc}
        \toprule
        $\theta^\mathrm{iou}$ & w/o & 0.2 & 0.4 & 0.6 & 0.8 \\
        \midrule
        Novel & 30.4 &30.5  &30.8  & 31.0 &30.4  \\
        \color{gray}{Base}  & \color{gray}{52.8} &\color{gray}{52.6}  &\color{gray}{52.6}  & \color{gray}{52.9} &\color{gray}{52.0}  \\
        % \color{gray}{All}   & \color{gray}{47.0} &\color{gray}{46.8}  &\color{gray}{46.8}  & \color{gray}{47.2} &\color{gray}{46.3}  \\
        \bottomrule
    \end{tabular}
    }
    \label{tab:ablation_on_IOU}
    \vspace{-1em}
    
\end{table}
\begin{table}[t]
    \centering
    \caption{The results of various $\gamma$ in OCA on COCO.}
    \footnotesize{
    \begin{tabular}{l|cccccc}
        \toprule
         & 0.0 & 0.2 & 0.4 & 0.6 & 0.8 & 1.0 \\
        \midrule
        Novel & 32.0 & 32.5 & 32.6 & 32.6 & 32.6 & 32.4 \\
        \color{gray}{Base}  & \color{gray}{53.1} & \color{gray}{53.3} & \color{gray}{53.4} & \color{gray}{53.5} & \color{gray}{53.4} & \color{gray}{53.3} \\
        \bottomrule
    \end{tabular}
    }
    \label{tab:effect_of_gamma_in_CBA}
\end{table}

\section{Conclusion}
% In this paper, we present MEDet, a novel and effective framework to solve the two observable challenges for better OVD performance.
% To sharpen the visual-language knowledge of pre-trained models from image-level to proposal-level, we devise an online proposal mining scheme. Moreover, to generate accurate predictions on both base categories and novel ones without bias, we propose a class-wise backdoor adjustment.
% Experiments on popular OVD benchmarks demonstrate that the proposed method outperforms the counterpart approaches in detecting objects of novel categories.

In this paper, we present MEDet, a novel and effective framework to solve the two observable challenges for better OVD results.
To sharpen the vision-language knowledge of caption datasets from image-level to proposal-level and learn an unbounded vocabulary of concepts, we devise an online proposal mining scheme in an end-to-end network to fully explore the knowledge of caption data. Moreover, we propose an offline class-wise adjustment to generate accurate predictions on both base and novel categories with less bias.
Experiments on popular COCO and LVIS benchmarks verify that our MEDet outperforms the counterpart approaches in detecting objects of novel categories.
% For future work, we can design powerful detector or optimize a stronger RPN to generate higher quality for vision-language pairs to further improve the generalization of new categories.

%%%%%%%%% REFERENCES
{\small
\bibliographystyle{ieee_fullname}
\bibliography{PaperForReview}
}

\end{document}